\documentclass[letterpaper]{article} 
\usepackage{aaai2027}  
\usepackage{amsmath,amssymb}
\usepackage[hyphens]{url}  
\usepackage{graphicx} 
\usepackage{natbib}  
\usepackage{caption} 
\allowdisplaybreaks
\usepackage{algorithm}
\usepackage{algorithmic}

\usepackage{newfloat}
\usepackage{listings}
\DeclareCaptionStyle{ruled}{labelfont=normalfont,labelsep=colon,strut=off} 
\floatstyle{ruled}
\newfloat{listing}{tb}{lst}{}
\floatname{listing}{Listing}

\usepackage{booktabs}
\usepackage{multirow}
\usepackage{colortbl}
\usepackage{tabularx}

\newcommand{\method}{EmoTrace}

\title{EmoTrace: Recursive Trace-Factorized Generation for Emotional Support Conversation}
\author{
    Mingwen Zhang \textsuperscript{\rm 1},
    Minqiang Yang \textsuperscript{\rm 1, *},
    Yang Yu \textsuperscript{\rm 1},
    Haobo Liang \textsuperscript{\rm 1},
    Yongfeng Tao \textsuperscript{\rm 1},
    Changsheng Ma \textsuperscript{\rm 1, *},
    Bin Hu \textsuperscript{\rm 1, *},
}
\affiliations{
    \textsuperscript{\rm 1}School of Information Science and Engineering, Lanzhou University, Lanzhou, China\\

    \{zhangmw2024\}@lzu.edu.cn
}

\begin{document}

\maketitle

\begin{abstract}
Fundamental to effective emotional support is the iterative interpersonal dynamic wherein a supporter intervention continuously shapes the psychological trajectory of the seeker. While contemporary models capture variables at the turn level, including transient emotions and discrete strategies, they inherently treat these factors as isolated events. By retaining strictly literal dialogue histories while discarding the underlying decision logic, such architectures remain fundamentally incapable of sustaining coherent extended support. 
Here we introduce EmoTrace, a recursive method utilizing trace factorization to preserve explicit decision continuity throughout interactions spanning multiple turns.
Departing from others, our approach formulates each conversational turn as a structured interaction increment encapsulating the evolving state of the seeker, the selected strategy, the affect of the supporter, and the ultimate textual realization. 
Accumulated sequentially, this trajectory of increments functions as an explicit structural trace informing subsequent interventions. 
Optimization of this factorized generation is achieved through two complementary mechanisms. Specifically, a value guidance mechanism aligns discrete factor selection with interaction values over an extended horizon, whereas a trace conditioning mechanism refines textual realization contingent upon the fixed strategic trace. 
Empirical evaluations across standard benchmarks demonstrate that EmoTrace yields substantial improvements in both strategic planning and response generation. Crucially, rollouts spanning multiple turns reveal that these retained structural traces confer compounding benefits over extended dialogues, establishing a robust mechanism for sustained affective alignment in human-machine interaction.
\end{abstract}


\section{Introduction}

Modern Emotional Support Conversation architectures increasingly factorize the intervention process into discrete cognitive and affective stages, 
sequentially deducing seeker state, selecting strategic paradigms, and adopting empathetic stances prior to textual realization. 
While sequential extensions of this paradigm seek to modulate immediate interventions via projected future utility~\cite{liu2021esconv,tu2022misc,zhao2023transesc,cheng2022multiesc,zhou2023supporter,wang2025straq}, 
such explicit reasoning remains strictly circumscribed by the isolated turn boundary. 
Following textual generation, the underlying decision rationale effectively dissipates. 
Subsequent transitions are conditioned exclusively upon surface lexical histories, entirely devoid of the specific latent variables that orchestrated the preceding response. 
This architectural constraint engenders a profound asymmetry, 
rendering current models highly explicit within isolated conversational epochs yet fundamentally implicit across the broader interaction trajectory. 
Such structural discontinuity necessitates a critical reevaluation of state propagation, 
prompting a central inquiry regarding the precise internal representations an affective agent must preserve to sustain coherent longitudinal support.

\begin{figure}[t]
    \centering
    \includegraphics[width=\columnwidth]{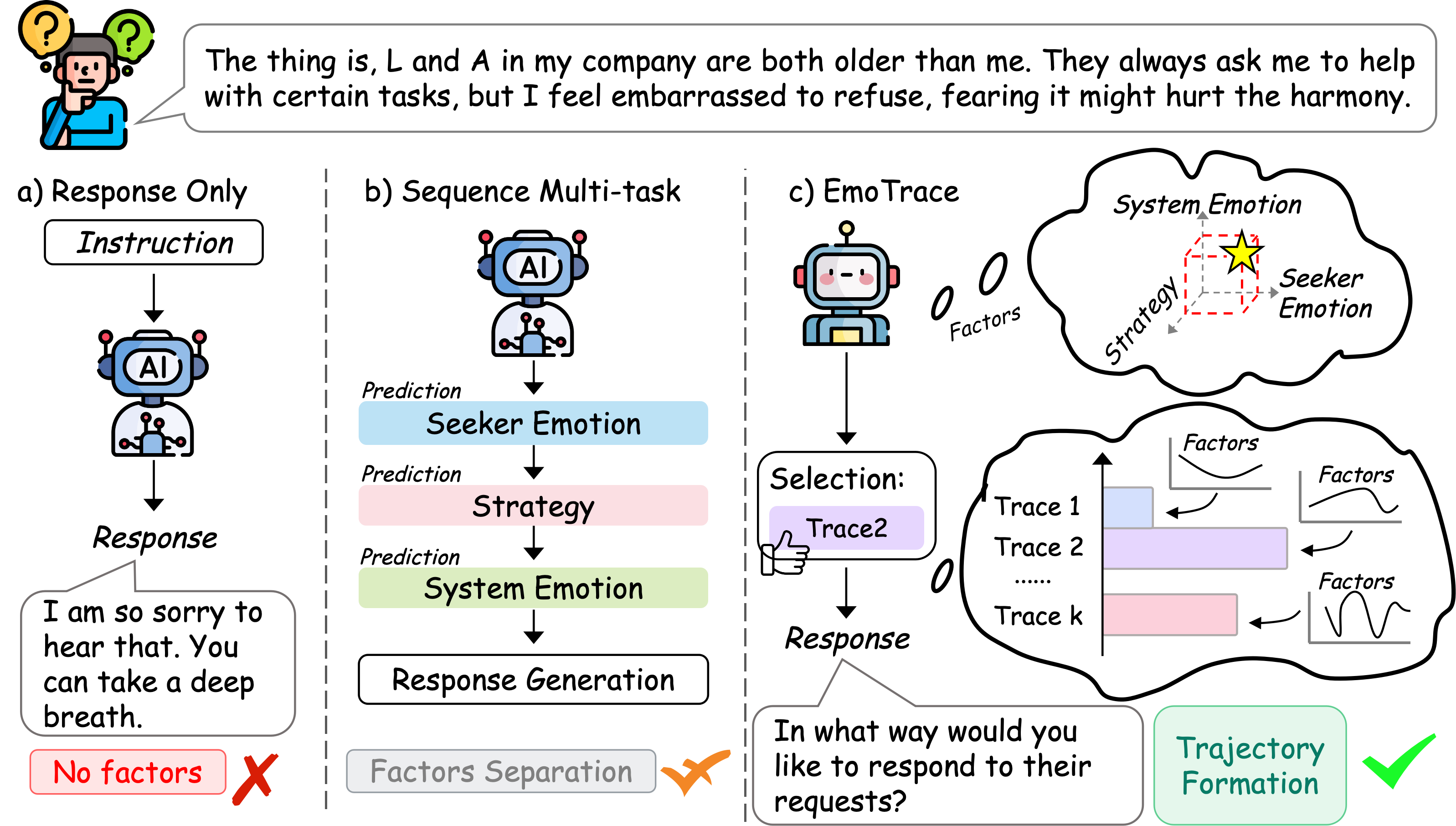}
    \caption{Three ESC generation paradigms: response-only, turn-local sequential
    factor generation, and recursive trace factorization. }
    \label{fig:interaction_landscape}
\end{figure}

The necessity for such state propagation derives fundamentally from social emotion regulation~\cite{reeck2016socialregulation}, 
which conceptualize interpersonal affective exchange as an iterative and historically contingent process. 
Within this paradigm, a supporter continuously interprets the affective state of the seeker, executes a regulatory strategy, and assimilates the subsequent reaction as updated contextual evidence. 
They elucidate a critical dependency systematically overlooked by standard generation objectives, 
dictating that subsequent interventions are jointly conditioned upon incoming feedback and the prior mapping of cognitive interpretation to behavioral action. 
This architectural limitation is starkly exemplified by the MESC framework~\cite{chu2025mesc}. 
Although its sequential routine successfully factorizes user emotion, system strategy, and supporter affect prior to textual realization within an isolated turn, 
the ensuing conversational epochs retain exclusively the surface utterances and multimodal observations, thereby irrevocably discarding the underlying generative state. 
Addressing this deficit therefore transcends the mere incorporation of supplementary conversational features or discrete planning modules. 
It fundamentally necessitates a persistent internal trace, 
operating as a structural memory that intrinsically tethers latent state estimations and strategic deliberations to the precise textual realizations that instantiate them.

\method{} bridges this structural discontinuity by formulating each supporter intervention as a \textit{factorized interaction increment}, comprising a structured factor configuration encoding the evolving state of the seeker, the selected regulatory strategy, and the affective stance of the supporter, followed by a textual realization strictly conditioned upon that configuration. Appended sequentially to the interaction trajectory, each complete increment ensures that subsequent turns are jointly conditioned upon both emergent seeker input and the structural decisions instantiated earlier in the dialogue, with all components produced through a unified autoregressive pass.

Acquisition of this representation proceeds through two sequential optimization stages. \textit{Value Guided Trace Factorization} (VGTF) optimizes the full increment by steering the factor conditional toward configurations associated with elevated interaction value over extended horizons, enabling trajectory level feedback to directly modulate factor generation while obviating the need for a discrete planning module during inference. \textit{Trace Conditioned Relative Optimization} (TCRO) subsequently freezes the factor conditional and refines the response conditional in isolation, evaluating candidate responses under matched factor configurations against utility and feasibility costs derived from the conditioned trace. This bifurcated optimization decouples structural deliberation from linguistic instantiation, preserving strategic coherence while simultaneously maximizing expressive fidelity.

Empirical evaluations comprising sequential dialogic rollouts and blinded preference assessments across the MESC and ESConv benchmarks reveal a performance trajectory strictly contingent upon the interaction horizon. During brief exchanges, the proposed architecture yields marginal empirical advantages over established baselines. As conversational epochs extend and accumulated increments compound, however, this performance divergence expands substantially. Such scaling dynamics empirically corroborate our central hypothesis, demonstrating that a persistent internal trace effectively mitigates the structural asymmetry inherent in strictly localized conversational reasoning. Fundamentally, these findings establish that sustained affective support relies not merely upon superficial lexical histories, but rather upon conditioning subsequent interventions on an evolving repository of latent interpretive states. The principal contributions of this investigation encompass three fundamental dimensions:

\begin{itemize}
    \item \textbf{EmoTrace architecture.} We introduce EmoTrace, an architecture that formulates emotional support conversation as a recursive trace factorization process. 
    The model generates structured interaction increments comprising latent factor configurations and conditioned responses grounded in multimodal contexts. 
    \item \textbf{Value-Guided Trace Factorization (VGTF).} We develop VGTF to optimize structural decision-making by biasing factor generation toward configurations associated with high long-horizon interaction value. 
    \item \textbf{Trace-Conditioned Relative Optimization (TCRO).} We design TCRO to refine the linguistic realization of supporter responses by decoupling textual articulation from structural planning. 
\end{itemize}
 
\section{Related Work}

\subsection{Emotional Support Conversation}
Basic research in empathetic dialogue established response generation grounded in emotion, affect, and commonsense reasoning ~\cite{ma2020survey,rashkin2019empathetic,lin2019moel,majumder2020mime,sabour2022cem}. 
Subsequent investigations enriched this paradigm by incorporating persona ~\cite{zhong2020persona,cheng2023pal}, integrating emotional knowledge ~\cite{li2022kemp}, 
and combining mental health principles with mixed initiative interactions ~\cite{deng2023kemi}. 
Pivoting toward structured intervention, ESConv formalized explicit strategy selection, 
while MISC and TransESC coupled seeker state inference with localized semantic and affective transitions ~\cite{liu2021esconv,tu2022misc,zhao2023transesc}. 
Although feedback responsive and hierarchical models further aligned cognition with generated responses ~\cite{peng2023fado,zhou2023case}, these architectures predominantly optimize isolated conversational epochs. 
The MESC framework advances this trajectory by providing synchronized annotations for user emotion, system strategy, and supporter affect ~\cite{chu2025mesc}. 
Its sequential routine autoregressively instantiates these latent factors alongside the textual response within an isolated turn. 
Crucially, however, the ensuing dialogue history retains only the surface utterances, irrevocably discarding the underlying factor configuration. 

\begin{figure*}[t]
    \centering
    \includegraphics[width=0.95\textwidth]{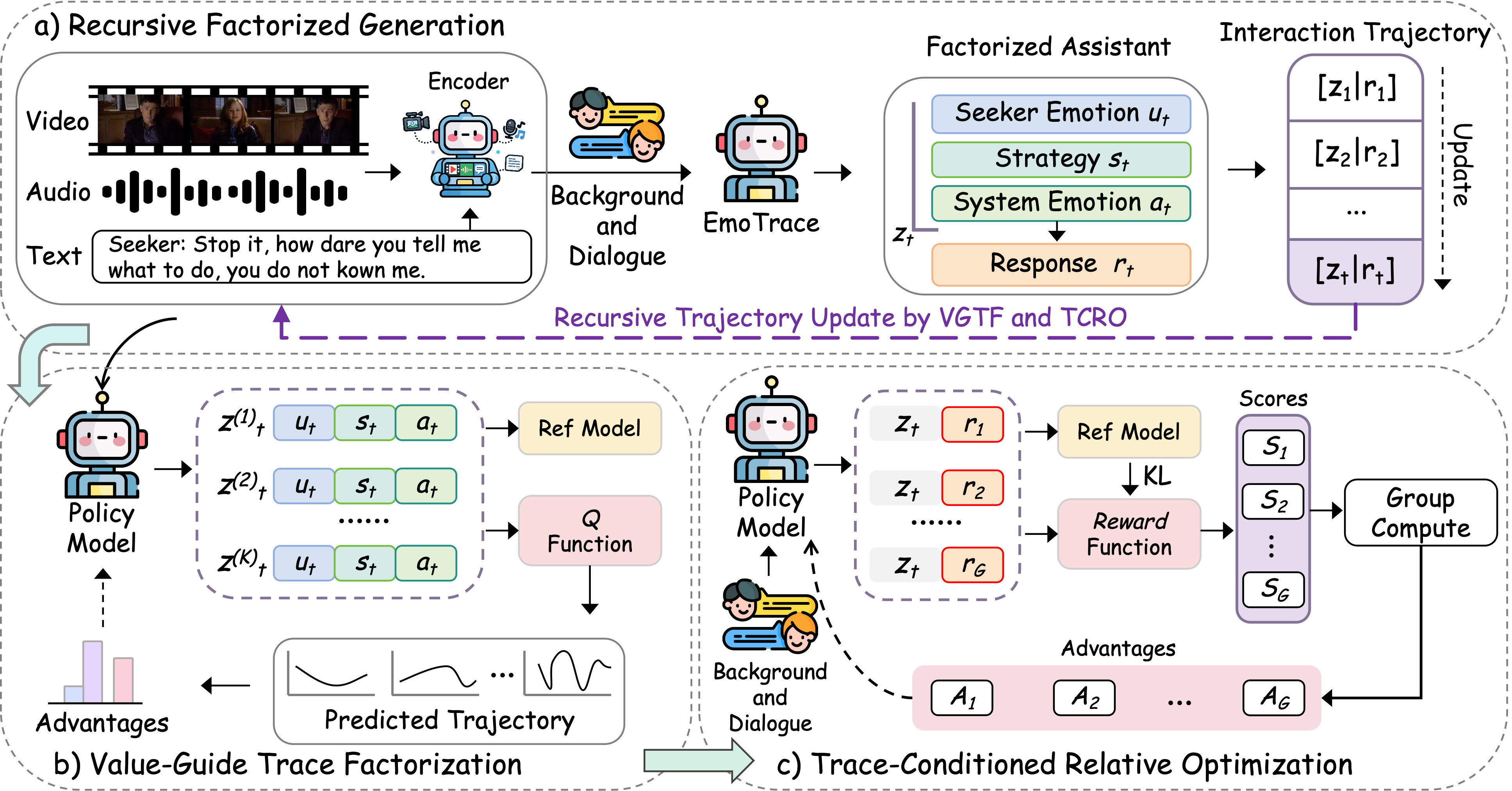}
    \caption{
    Two-stage training of \method{}.
    Video-LLaMA converts video and audio signals into textual nonverbal cues,
    which are combined with the current seeker utterance and the retained
    interaction trajectory.
    VGTF learns complete factorized increments and guides the factor
    conditional with long-horizon factor values.
    TCRO freezes the factor conditional and optimizes the response
    conditional through grouped response comparison under the same factor
    configuration.
    }
    \label{fig:framework}
\end{figure*}

\subsection{Long-Horizon Dialogue Planning}

Multi-turn ESC has motivated explicit support-strategy planning. MultiESC selects
strategies from lookahead estimates of seeker feedback
~\cite{cheng2022multiesc}; positive-emotion elicitation optimizes support over
multiple turns~\cite{zhou2023supporter}; Dynamic Strategy Prompt Reasoning
constructs adaptive prompts from inferred seeker intentions
~\cite{liu2025dynamicstrategy}; and straQ* uses Q-learning to estimate
long-horizon strategy value~\cite{wang2025straq}. ESCA combines a planner over
interaction variables with a reinforcement-learned, strategy-aligned prompt
generator~\cite{li2026esca}, while ProphetChat simulates future conversation
before generating the current response~\cite{liu2022prophetchat}. Related
dialogue models organize generation through interpretable discrete
representations~\cite{zhao2018discrete}, latent action spaces optimized by
reinforcement learning~\cite{zhao2019latentactions}, discrete latent acts
~\cite{bao2020plato}, or explicit policy injection~\cite{he2022galaxy}. These
methods use future estimates or intermediate actions primarily to control a
current strategy or response. 

\subsection{Reinforcement Learning for Response Generation}

Preference-based reinforcement learning first learned reward signals from human
comparisons~\cite{christiano2017humanpreferences}; language-model alignment then
used those signals with KL-regularized policy optimization
~\cite{stiennon2020summarize,ziegler2019finetuning}. Dialogue-specific inverse
reinforcement learning also treats response generation as policy learning from
demonstrations~\cite{li2019dialogueirl}. Proximal Policy Optimization (PPO) uses
clipped importance-ratio updates~\cite{schulman2017ppo}; Direct Preference
Optimization (DPO) converts pairwise preferences into a direct policy objective
~\cite{rafailov2023dpo}; and group-relative optimization forms critic-free
relative advantages from sampled output groups~\cite{shao2024deepseekmath}. In
ESC, DecoupledESC applies preference optimization to separate strategy planning
and response generation~\cite{zhang2025decoupledesc}, while multifaceted
feedback exposes response failures not captured by fluency alone
~\cite{wang2024muffin,kang2024preference}. 
However, these methods optimize response quality against generic preference signals without conditioning on a pre-committed structural trace. TCRO complements this line of work by constraining response optimization to remain faithful to factor configurations produced by VGTF, enabling separate governance of the seeker and supporter trajectories.
 
\section{Method}

\subsection{Overview}

\method{} represents each supporter turn as a factorized interaction increment
consisting of a structured factor configuration and its textual realization.
The generated increment is appended to the interaction trajectory and retained
as context for subsequent turns, yielding a recursive process of trajectory
construction (Figure~\ref{fig:framework}).


\subsection{Problem Formulation}

At supporter turn $t$, the observed input consists of the seeker utterance $U_t$, video $V_t$, and audio $A_t$. A frozen multimodal encoder (e.g., Video-LLaMA) maps the nonverbal channels into a textual cue sequence; together with the retained trajectory $\mathcal T_{<t}$, this yields the policy context:
\begin{equation}\label{eq:multimodal_encoding}
d_t^{\mathrm{AV}}=E_{\mathrm{VL}}(V_t,A_t),\qquad
\mathbf x_t=(U_t,d_t^{\mathrm{AV}},\mathcal T_{<t}).
\end{equation}

The factor configuration is $\mathbf z_t=(u_t,s_t,a_t)$, where $u_t$, $s_t$, and $a_t$ denote the seeker state, support strategy, and supporter affect, respectively, each represented as a categorical label. Concatenating this configuration with the response realization $r_t$ gives the complete interaction increment and the recursive trajectory update:
\begin{equation}\label{eq:interaction_increment}
\begin{aligned}
\boldsymbol\tau_t&=(\mathbf z_t,r_t),\qquad
\mathcal T_{\leq t}=\mathcal T_{<t}\oplus\boldsymbol\tau_t,\\
\mathbf x_{t+1}&=(U_{t+1},d_{t+1}^{\mathrm{AV}},\mathcal T_{\leq t}).
\end{aligned}
\end{equation}

Let $\mathbf y_t$ be the serialized token sequence of the increment, where each factor label is tokenized as text and concatenated with the response tokens. The joint distribution factorizes autoregressively in a single decoding pass:
\begin{equation}\label{eq:trace_autoregressive}
\begin{aligned}
p_\theta(\boldsymbol\tau_t\mid\mathbf x_t)
&=p_\theta^{F}(\mathbf z_t\mid\mathbf x_t)
  p_\theta^{R}(r_t\mid\mathbf x_t,\mathbf z_t)\\
&=\prod_{\ell \in \mathcal I_F}
p_\theta^{F}(y_{t,\ell}\mid\mathbf x_t,\mathbf y_{t,<\ell}) \\
&\quad \times \prod_{\ell \in \mathcal I_R}
p_\theta^{R}(y_{t,\ell}\mid\mathbf x_t,\mathbf y_{t,<\ell}),
\end{aligned}
\end{equation}
where $\mathcal I_F$ and $\mathcal I_R$ denote the disjoint token index spans for the factor configuration and the response, respectively. The conditionals $p_\theta^{F}$ and $p_\theta^{R}$ are computed as contiguous spans of a single decoder over a frozen backbone $\theta_0$, utilizing span-routed blocks $\eta_F$ and $\eta_R$ that specifically score the factor positions $\mathcal I_F$ and response positions $\mathcal I_R$.

\subsection{Value-Guided Trace Factorization}

VGTF learns the full interaction increment while biasing the factor
conditional toward configurations with higher extended horizon values. A localized judge $q_\xi$ evaluates the immediate quality of the observed increment. 
Assuming the judge yields a bounded score with a maximum value $M$, 
the normalized local reward is defined as
\begin{equation}\label{eq:vgtf_base_reward}
b_t
=
\frac{1}{M}
q_\xi
\left(
\mathbf x_t,
\mathbf z_t,
r_t,
\mathbf x_{t+1}
\right).
\end{equation}

A trajectory-prefix potential $\Phi_\omega$ maps the current interaction
context to a scalar preference potential. The shaped reward is
\begin{equation}\label{eq:vgtf_shaped_reward}
\widetilde b_t
=
b_t
+
\gamma
\Phi_\omega(\mathbf x_{t+1})
-
\Phi_\omega(\mathbf x_t),
\end{equation}
where $\gamma$ is the discount factor.

We define the factor value as the long-horizon utility associated with
selecting configuration $\mathbf z_t$ under context $\mathbf x_t$ over the
remaining $H{-}t$ turns of the trajectory:
\begin{equation}\label{eq:factor_value}
Q_F(\mathbf x_t,\mathbf z_t)
=
\mathbb E
\left[
\sum_{k=0}^{H-t}
\gamma^k
\widetilde b_{t+k}
\,\middle|\,
\mathbf x_t,\mathbf z_t
\right].
\end{equation}
A value estimator $Q_\phi$ is trained to approximate $Q_F$; its training
objective is detailed in the Supplement.

The estimated factor value induces a value-guided factor distribution:
\begin{equation}\label{eq:value_guided_factor_policy}
\pi_V(\mathbf z_t\mid\mathbf x_t)
=
\frac{
p_0^F(\mathbf z_t\mid\mathbf x_t)
\exp
\left(
Q_\phi(\mathbf x_t,\mathbf z_t)/\tau_F
\right)
}{
Z(\mathbf x_t)
},
\end{equation}
where $p_0^F$ is the supervised factor prior, $\tau_F$ is the value
temperature, and $Z(\mathbf x_t)$ is the normalizing constant.

The complete-increment likelihood is
\begin{equation}\label{eq:increment_likelihood}
\mathcal L_{\mathrm{inc}}
=
-
\mathbb E_{(\mathbf x,\boldsymbol\tau)\sim\mathcal D}
\log
p_\theta(\boldsymbol\tau\mid\mathbf x).
\end{equation}
VGTF minimizes the following objective:
\begin{equation}\label{eq:vgtf_objective}
\mathcal L_{\mathrm{VGTF}}
=
\mathcal L_{\mathrm{inc}}
+
\alpha_F
\mathbb E_{\mathbf x\sim\mathcal D}
\mathbb E_{\mathbf z \sim \pi_V(\cdot\mid\mathbf x)}
\left[
-
\log
p_\theta^F(\mathbf z\mid\mathbf x)
\right].
\end{equation}
The likelihood term trains both factor and response spans, while the
second term acts as a value-guided projection, optimizing only the factor span via behavior cloning towards the target distribution $\pi_V$.

\subsection{Trace-Conditioned Relative Optimization}

TCRO fixes the factor conditional parameters at their VGTF-optimized values and subsequently optimizes only the response conditional. 
For each context $\mathbf x_t$, a factor configuration is first sampled from the fixed factor policy:
\begin{equation}\label{eq:tcro_factor_sample}
\widehat{\mathbf z}_t
\sim
p_\theta^F(\cdot\mid\mathbf x_t).
\end{equation}
The response policy is then optimized under the trace-conditioned context
\begin{equation}\label{eq:tcro_context}
\mathbf h_t
=
(\mathbf x_t,\widehat{\mathbf z}_t).
\end{equation}

A group of $G$ candidate responses is sampled from the old response policy:
\begin{equation}\label{eq:tcro_group}
\mathcal G_t
=
\{r_{t,i}\}_{i=1}^{G},
\qquad
r_{t,i}
\sim
p_{\mathrm{old}}^R(\cdot\mid\mathbf h_t).
\end{equation}

Each candidate is assigned a constrained score:
\begin{equation}\label{eq:tcro_score}
S_{t,i}
=
U_\psi(r_{t,i}\mid\mathbf h_t)
-
\sum_{j=1}^{J}
\lambda_j
\left[
C_j(r_{t,i};\mathbf h_t)-\varepsilon_j
\right]_+,
\end{equation}
where $U_\psi$ is the factor-conditioned response utility, $C_j$ is the
$j$-th constraint cost, $\varepsilon_j$ is its budget, and $\lambda_j$ is its
dual weight. Constraints are enforced softly through this penalization: the dual weights $\lambda_j$ adapt online so that violations approaching their budgets are penalized more sharply, while feasible candidates receive no penalty.

The group-relative advantage is computed from the standardized constrained
scores:
\begin{equation}\label{eq:tcro_advantage}
A_{t,i}
=
\frac{
S_{t,i}-\mu_t
}{
\sigma_t+\epsilon_A
}
+
\kappa
\left(
\mathbb I[i=i_t^\star]
-
\frac{1}{G}
\right)m_t,
\end{equation}
where the mean and standard deviation of the group scores are
\begin{equation}\label{eq:tcro_score_stats}
\mu_t
=
\frac{1}{G}
\sum_{i=1}^{G}S_{t,i},
\qquad
\sigma_t
=
\sqrt{
\frac{1}{G}
\sum_{i=1}^{G}
(S_{t,i}-\mu_t)^2
},
\end{equation}
and the winner index $i_t^\star$ and margin term $m_t$ are defined in the
Supplement.

For response token $k\in\mathcal I_R$ of candidate $i$, the policy ratio is
\begin{equation}\label{eq:tcro_ratio}
\rho_{t,i,k}
=
\frac{
p_\theta^R
\left(
r_{t,i,k}
\mid
\mathbf h_t,r_{t,i,<k}
\right)
}{
p_{\mathrm{old}}^R
\left(
r_{t,i,k}
\mid
\mathbf h_t,r_{t,i,<k}
\right)
}.
\end{equation}
Its clipped version is
\begin{equation}\label{eq:tcro_clipped_ratio}
\bar\rho_{t,i,k}
=
\operatorname{clip}
\left(
\rho_{t,i,k},
1-\epsilon,
1+\epsilon
\right).
\end{equation}
Clipping is applied per token because the response policy is updated while the factor policy remains frozen; bounding each token-level ratio prevents large response-side shifts from desynchronizing with the committed factor configuration.

Following clipped group-relative policy optimization, TCRO minimizes the following objective:
\begin{equation}\label{eq:tcro_objective}
\resizebox{\columnwidth}{!}{%
$\displaystyle
\begin{aligned}
\mathcal L_{\mathrm{TCRO}}
={}& - \mathbb E_{\substack{t \\ 1\leq i\leq G}} \bigg[ \frac{1}{|\mathcal I_R|} \sum_{k\in\mathcal I_R} \min \Big( \rho_{t,i,k}A_{t,i},\, \bar\rho_{t,i,k}A_{t,i} \Big) \bigg] \\
&+ \beta \mathbb E_{\substack{t \\ 1\leq i\leq G}} \bigg[ \frac{1}{|\mathcal I_R|} \sum_{k\in\mathcal I_R} D_{\mathrm{KL}} \Big( p_\theta^R(\cdot|\mathbf h_t,r_{t,i,<k}) \,\big\|\, p_{\mathrm{ref}}^R(\cdot|\mathbf h_t,r_{t,i,<k}) \Big) \bigg]
\end{aligned}$%
}
\end{equation}
Here, $p_{\mathrm{ref}}^R$ is the supervised response reference policy. 
Only the response span parameters $\eta_R$ are optimized during this stage.

\begin{table*}[t]
\centering
\small
\renewcommand{\arraystretch}{1.06}
\setlength{\tabcolsep}{1.0pt}
\begin{tabular*}{\textwidth}{@{\extracolsep{\fill}}l*{9}{c}@{}}
\toprule
& \multicolumn{6}{c}{\bfseries Factor Prediction}
& \multicolumn{3}{c}{\bfseries Response Realization} \\
\cmidrule(lr){2-7}\cmidrule(l){8-10}
& \multicolumn{2}{c}{\bfseries User Emotion}
& \multicolumn{2}{c}{\bfseries System Strategy}
& \multicolumn{2}{c}{\bfseries System Emotion}
& & & \\
\cmidrule(lr){2-3}\cmidrule(lr){4-5}\cmidrule(lr){6-7}
\bfseries Model
& \bfseries Acc.$\uparrow$ & \bfseries W-F1$\uparrow$
& \bfseries Acc.$\uparrow$ & \bfseries W-F1$\uparrow$
& \bfseries Acc.$\uparrow$ & \bfseries W-F1$\uparrow$
& \bfseries B-2$\uparrow$ & \bfseries R-L$\uparrow$ & \bfseries BS-F1$\uparrow$ \\
\midrule
\multicolumn{10}{@{}l}{\bfseries Benchmarks} \\
DialogueGCN      & 46.27 & 50.61 & --    & --    & --    & --    & --   & --    & --    \\
MMGCN            & 55.80 & 57.58 & --    & --    & --    & --    & --   & --    & --    \\
MMDFN            & 58.13 & 55.86 & --    & --    & --    & --    & --   & --    & --    \\
\shortstack[l]{BlenderBot-\\Joint} & --    & --    & 48.00 & 46.10 & --    & --    & 4.85 & 15.25 & 85.50 \\
BBMHR            & --    & --    & --    & --    & --    & --    & 1.31 & 15.38 & 86.60 \\
GPT-3.5          & 33.50 & 33.80 & 19.90 & 17.60 & 17.40 & 27.60 & 1.01 & 4.60  & 84.50 \\
GPT-4            & 15.73 & 15.80 & 9.73  & 11.25 & 14.04 & 14.09 & 4.98 & 9.96  & 84.60 \\
SMES             & 54.60 & 46.80 & 49.00 & 20.20 & 96.10 & 64.00 & 5.13 & 15.42 & 86.80 \\
\midrule
\multicolumn{10}{@{}l}{\bfseries Controlled Methods} \\
Direct Response
& -- & -- & -- & -- & -- & --
& 4.70$\pm$0.08 & 14.80$\pm$0.15 & 85.90$\pm$0.06 \\
\shortstack[l]{Input-Factor\\Prompt}
& 80.00$\pm$0.24 & 79.71$\pm$0.28
& 58.50$\pm$0.31 & 57.69$\pm$0.34
& 99.30$\pm$0.05 & 99.36$\pm$0.06
& 5.00$\pm$0.09 & 15.60$\pm$0.19 & 86.20$\pm$0.07 \\
Serial Cascade
& 83.04$\pm$0.18 & 82.74$\pm$0.21
& 61.70$\pm$0.26 & \underline{61.48}$\pm$0.29
& \underline{99.72}$\pm$0.02 & \underline{99.72}$\pm$0.02
& 4.16$\pm$0.10 & 13.97$\pm$0.17 & 85.73$\pm$0.08 \\
Trace SFT
& 83.60$\pm$0.16 & 83.36$\pm$0.19
& 61.70$\pm$0.22 & 60.98$\pm$0.25
& 99.50$\pm$0.02 & 99.63$\pm$0.03
& 5.70$\pm$0.09 & 17.40$\pm$0.16 & 86.50$\pm$0.05 \\
straQ*
& \underline{83.68}$\pm$0.17 & \underline{83.45}$\pm$0.20
& \underline{62.12}$\pm$0.25 & 61.45$\pm$0.28
& 99.63$\pm$0.02 & 99.63$\pm$0.02
& 5.68$\pm$0.07 & 17.92$\pm$0.13 & 86.66$\pm$0.05 \\
DecoupledESC
& 83.21$\pm$0.19 & 82.98$\pm$0.22
& 61.84$\pm$0.24 & 61.17$\pm$0.27
& 99.61$\pm$0.03 & 99.61$\pm$0.03
& \underline{5.94}$\pm$0.08 & \underline{18.20}$\pm$0.12 & \underline{86.75}$\pm$0.04 \\
\midrule
\rowcolor{gray!10}\textbf{EmoTrace}
& \bfseries 85.18$\pm$0.15\textsuperscript{$\dagger$} 
& \bfseries 84.95$\pm$0.17\textsuperscript{$\dagger$}
& \bfseries 64.25$\pm$0.20\textsuperscript{$\dagger$} 
& \bfseries 63.71$\pm$0.23\textsuperscript{$\dagger$}
& \bfseries 99.75$\pm$0.02\textsuperscript{$\dagger$}
& \bfseries 99.75$\pm$0.01\textsuperscript{$\dagger$}
& \bfseries 6.24$\pm$0.07\textsuperscript{$\dagger$}
& \bfseries 19.21$\pm$0.11\textsuperscript{$\dagger$}
& \bfseries 87.12$\pm$0.04\textsuperscript{$\dagger$} \\
\bottomrule
\end{tabular*}
\caption{MESC factor prediction and response realization. -- marks an unevaluated task. 
Controlled Accuracy, Weighted-F1, and response metrics report five-seed mean${}\pm{}$SD. $\dagger$ denotes $p<0.05$ against the strongest eligible five-seed baseline.}
\label{tab:mesc_main}
\end{table*}

\section{Experiments}

\subsection{Experimental Setup}

\noindent\textbf{Datasets.}
We evaluate our method on two widely-used Emotional Support Conversation (ESC) benchmarks: MESC~\cite{chu2025mesc} and ESConv~\cite{liu2021esconv}. MESC provides fine-grained annotations for user emotion, system strategy, and system emotion. Following its standard protocol, we use 815/102/102 dialogues (comprising 8,252/1,001/1,073 response instances) for the training, validation, and test sets, respectively. ESConv provides supervision for seeker states and support strategies; we adopt its standard dialogue-level split of 632/210/211.

\noindent\textbf{Baselines.}
The MESC comparison reproduces the benchmark rows reported in its source table~\cite{chu2025mesc}: DialogueGCN~\cite{ghosal2019dialoguegcn}, MMGCN~\cite{hu2021mmgcn}, MM-DFN~\cite{hu2022mmdfn}, BlenderBot-Joint~\cite{roller2021blenderbot}, BBMHR, GPT-3.5, GPT-4~\cite{openai2023gpt4}, and SMES. Same-backbone controls comprise Direct Response, Input-Factor Prompt, Serial Cascade, and Trace SFT, together with MESC implementations of MultiESC~\cite{cheng2022multiesc}, straQ*~\cite{wang2025straq}, and DecoupledESC~\cite{zhang2025decoupledesc}. Trace SFT applies supervised fine-tuning to the complete factorized interaction increment without VGTF value guidance or TCRO response-policy optimization. 

\noindent\textbf{Evaluation Metrics.}
Our evaluation spans three complementary dimensions.

\textit{Classification and Attribution.} We report Accuracy and Weighted-F1 for the overall MESC task, and Macro-F1 for ESConv strategy prediction and VGTF factor attribution.

\textit{Response Generation.} Generation quality is measured via BLEU-2~\cite{papineni2002bleu}, ROUGE-L~\cite{lin2004rouge}, and BERTScore-F1~\cite{zhang2020bertscore}. On ESConv, we additionally report the dataset-native helpfulness and emotion-intensity scores.

\textit{Multi-Turn Evaluation.} Long-term effectiveness is quantified by the realized trajectory return $J_H=\sum_{k=0}^{H-1}\gamma^k b_{t+k}$ at rollout horizons $H\in\{3,5,8\}$. We further conduct blinded human evaluations over complete $H{=}8$ interactions to capture trajectory-level preference, and separately assess TCRO response preference by comparing immediate outputs under a fixed factor configuration.


\noindent\textbf{Implementation Details.}
All same-backbone controlled systems use Qwen2.5-3B-Instruct~\cite{yang2024qwen25}, LoRA~\cite{hu2022lora}, an eight-turn history window, 1,024 input tokens, 192 output tokens, and seeds $\{13,21,42,87,101\}$. The multi-turn study uses fixed Qwen2.5-3B-Instruct and Llama-3.2-3B-Instruct seeker simulators from their respective model families~\cite{grattafiori2024llama3}, initialized from shared dialogue contexts. The supplement details the comparison, training, and rollout protocols.

\subsection{Factor Prediction and Response Quality}

Table~\ref{tab:mesc_main} reveals that general-purpose models and sequential pipelines struggle with empathetic dialogue due to insufficient implicit reasoning and severe error cascading. Coupling factor prediction with text generation in Trace SFT mitigates these cascading failures to establish a strong baseline for local lexical fluency. EmoTrace advances this coupled architecture by introducing value guidance to optimize long-term conversational returns rather than mere local accuracy. This mechanism substantially improves system strategy prediction while system emotion accuracy remains saturated by the underlying dataset distribution. Crucially, the resulting generation gains exhibit a distinct metric asymmetry. EmoTrace drives a substantial increase in deep semantic alignment alongside marginal improvements in surface n-gram overlap. This divergence demonstrates that factorized trajectory modeling primarily governs global content selection rather than superficial phrasing. Structural trajectory modeling thus proves highly effective at determining the appropriate empathetic intent rather than merely refining its linguistic realization.

\subsection{Cross-Dataset Generalization}

\begin{table}[t]
\centering
\small
\renewcommand{\arraystretch}{1.10}
\setlength{\tabcolsep}{1.5pt}
\begin{tabular*}{\columnwidth}{@{\extracolsep{\fill}}l*{6}{c}@{}}
\toprule
\bfseries Model
& \bfseries Se.$\uparrow$
& \bfseries St.$\uparrow$
& \bfseries B-2$\uparrow$
& \bfseries R-L$\uparrow$
& \bfseries Help.$\uparrow$
& \bfseries $\Delta$EI$\uparrow$ \\
\midrule
\multicolumn{7}{@{}l}{\bfseries Benchmarks} \\
Qwen, beam & -- & -- & 7.21 & 18.75 & -- & -- \\
\midrule
\multicolumn{7}{@{}l}{\bfseries Controlled Methods} \\
Trace SFT       & 72.10 & 39.24 & 7.12
                & 18.90 & 0.562 & 0.318 \\
MultiESC        & 72.84 & 40.78 & 7.18
                & 19.04 & 0.581 & 0.336\\
straQ*          & \underline{73.16} & \underline{41.35} & 7.24
                & 19.11 & 0.596 & 0.348 \\
DecoupledESC    & 72.92 & 40.96 & \underline{7.39}
                & \underline{19.28} & \underline{0.602} & \underline{0.351} \\
\midrule
\rowcolor{gray!10}\textbf{EmoTrace}
                & \bfseries 74.36
                & \bfseries 42.73
                & \bfseries 7.61
                & \bfseries 19.68
                & \bfseries 0.641
                & \bfseries 0.397\\
\bottomrule
\end{tabular*}
\caption{ESConv native-schema results. Se.\ and St.\ denote seeker-state and support-strategy Macro-F1; Help.\ and $\Delta$EI denote helpfulness and emotion-intensity improvement.}
\label{tab:esconv_replication}
\end{table}

\begin{figure}[t]
    \centering
    \includegraphics[width=0.9\columnwidth]{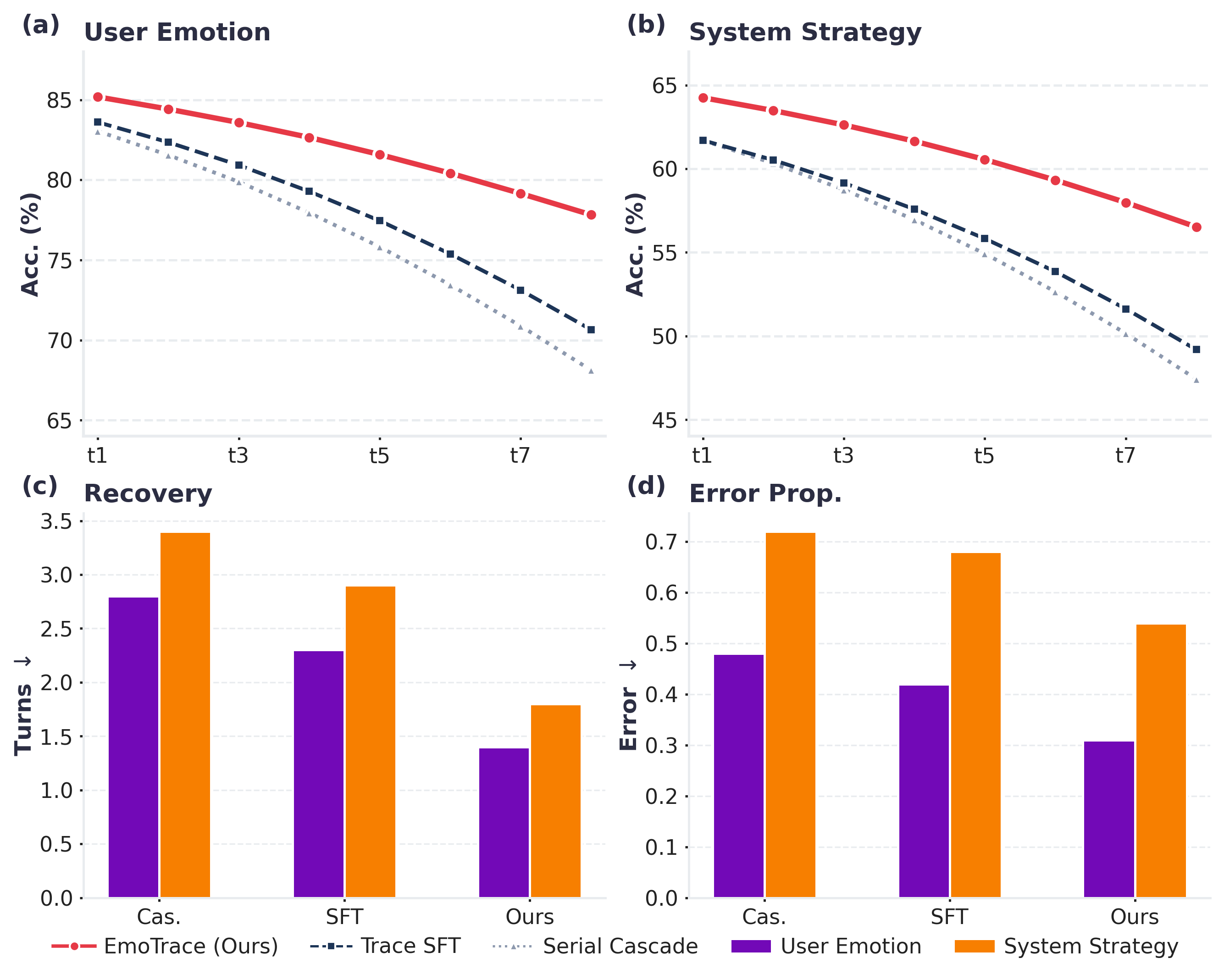}
    \caption{Multi-turn dynamics on MESC. (a) and (b) show per-turn factor accuracy. (c) shows average turns to recover from an error (lower is better). (d) shows conditional error propagation rate (lower is better).}
    \label{fig:additional_analysis}
\end{figure}

Table~\ref{tab:esconv_replication} demonstrates that the structural advantages of EmoTrace generalize robustly across distinct factor schemas. Evaluated on ESConv, which entirely eliminates the system-emotion dimension, the model maintains its performance progression by consistently outperforming baseline architectures in both factor prediction and text generation. This cross-schema robustness confirms that value-guided trajectory filtering optimizes structural reasoning without overfitting to specific emotional ontologies. Crucially, these architectural benefits translate directly into tangible interaction outcomes. By evaluating actual psychological impact rather than mere lexical overlap, native helpfulness and emotion-intensity improvement metrics reveal a substantial performance leap for EmoTrace. The model leverages its structured trajectory to select support strategies that genuinely address the cognitive distress of the seeker. Ultimately, these findings confirm that embedding latent conversational factors within the generation target fundamentally enhances dialogue utility rather than serving merely as an auxiliary prediction mechanism.
\subsection{Multi-turn Dynamics and Trajectory Evaluation}
\subsubsection{Multi-turn Interaction Quality}

Static single-turn metrics mask how retained increments affect downstream dynamics. Evaluating trajectory returns across horizons reveals that while EmoTrace matches the strongest comparator at $H=3$, its advantage widens significantly at $H=5$ and $H=8$, as detailed in Supplementary Table~S2. This horizon-dependent divergence confirms that retention yields cumulative long-term benefits rather than a uniform single-turn offset. Blinded human evaluations corroborate this finding. Across 240 paired trajectories at $H=8$, raters consistently preferred EmoTrace over all baselines, with detailed preference counts provided in Supplementary Table~S2.

\begin{table}[t]
\centering
\small
\renewcommand{\arraystretch}{1.04}
\setlength{\tabcolsep}{1.0pt}
\begin{tabular*}{\columnwidth}{@{\extracolsep{\fill}}l*{5}{c}@{}}
\toprule
\multicolumn{6}{@{}l}{\textbf{A.\ History Window Ablation}} \\
\midrule
\textbf{Window}
& \textbf{UE Acc}
& \textbf{SS Acc}
& \textbf{B-2}
& \textbf{R-L}
& \textbf{BS-F1} \\
\midrule
$K\!=\!2$  & 84.12 & 62.80 & 5.82 & 18.40 & 86.68 \\
$K\!=\!4$  & 84.58 & 63.35 & 5.98 & 18.72 & 86.85 \\
$K\!=\!6$  & 84.89 & 63.78 & 6.11 & 18.96 & 86.94 \\
\textbf{$K\!=\!8$}
           & \textbf{85.18} & \textbf{64.25} & \textbf{6.24} & \textbf{19.21} & \textbf{87.12} \\
\midrule
\multicolumn{6}{@{}l}{\textbf{B.\ Trajectory Quality}} \\
\midrule
\textbf{Model}
& \multicolumn{2}{c}{\textbf{Entropy$\uparrow$}}
& \multicolumn{3}{c}{\textbf{Judge$\uparrow$}} \\
\cmidrule(lr){2-3}\cmidrule(lr){4-6}
Serial Cascade    & \multicolumn{2}{c}{4.15} & \multicolumn{3}{c}{3.67} \\
Trace SFT         & \multicolumn{2}{c}{4.31} & \multicolumn{3}{c}{3.84} \\
\rowcolor{gray!10}\textbf{EmoTrace}
                  & \multicolumn{2}{c}{\textbf{4.82}} & \multicolumn{3}{c}{\textbf{4.21}} \\
\bottomrule
\end{tabular*}

\caption{Additional analysis on MESC. (A)~Factor-history window $K$ vs.\ factor prediction and response metrics. (B)~Trajectory diversity (sequence entropy) and internal consistency (LLM judge, 1--5). UE: User Emotion; SS: System Strategy.}
\label{tab:additional_analysis}
\end{table}

\begin{table*}[t]
\centering
\small
\renewcommand{\arraystretch}{1.04}
\setlength{\tabcolsep}{1.5pt}
\begin{tabular*}{0.95\textwidth}{@{\extracolsep{\fill}}l*{6}{c}@{}}
\toprule
\bfseries Configuration
& \bfseries \shortstack{User Emotion\\M-F1 $\uparrow$}
& \bfseries \shortstack{System Strategy\\M-F1 $\uparrow$}
& \bfseries BLEU-2 $\uparrow$
& \bfseries ROUGE-L $\uparrow$
& \bfseries BERTScore-F1 $\uparrow$
& \bfseries $J_8$ $\uparrow$ \\
\midrule
\rowcolor{gray!10}\textbf{EmoTrace (Full)}
& \textbf{85.05} & \textbf{64.20} & \textbf{6.24} & \textbf{19.21} & \textbf{87.12} & \textbf{2.64} \\
$-$ w/o VGTF
& 83.60 & 61.70 & 5.95 & 18.48 & 86.70 & 1.98 \\
$-$ w/o TCRO
& 85.05 & 64.20 & 5.86 & 18.31 & 86.77 & 2.64 \\
$-$ w/o Factor Retention
& 84.32 & 62.80 & 6.12 & 18.95 & 87.00 & 2.21 \\
$-$ w/o Trace Factorization
& $-$ & $-$ & 5.02 & 14.50 & 84.80 & 0.92 \\
\bottomrule
\end{tabular*}

\caption{Ablation of \method{} components. Each row removes one component from the full model. ``$-$'' marks inapplicable settings.}
\label{tab:ablation}
\end{table*}

\subsubsection{Multi-turn Dynamics}
\label{sec:dynamics}

Figure~\ref{fig:additional_analysis} dissects the mechanics of this multi-turn advantage. Panel a and Panel b show that baseline models like Trace SFT and Serial Cascade suffer monotonic accuracy decay from turn 1 to turn 8 due to error propagation through lexical history. EmoTrace exhibits significantly slower decay, outperforming Trace SFT by 7.3 points in system-strategy accuracy at turn 8. This resilience stems from grounding each turn in explicit prior factor commitments rather than noisy surface history.

This retention mechanism actively suppresses error cascades and accelerates correction. Panel c shows EmoTrace recovers from strategy errors in 1.8 turns and emotion errors in 1.4 turns. This vastly outpaces Trace SFT, which requires 2.9 and 2.3 turns respectively, as well as Serial Cascade at 3.4 and 2.8 turns. Consequently, the conditional error propagation rate drops sharply in Panel d. A strategy error under Trace SFT leads to a 68\% subsequent error rate, which EmoTrace reduces to 54\%. The user-emotion cascade rate similarly falls from 42\% to 31\%. Thus, retained configurations act as a stabilizing anchor that halts error cascading. Per-turn accuracy values are reported in Supplementary Table~S5.

\subsubsection{History Length and Trajectory Quality}

Table~\ref{tab:additional_analysis} further validates our architectural choices. Block A varies the factor-history window $K$ from 2 to 8. User-emotion accuracy improves monotonically from 84.12 to 85.18, while ROUGE-L increases from 18.40 to 19.21. This confirms that deeper explicit history yields cumulative benefits. Block B evaluates trajectory quality. EmoTrace achieves a sequence entropy of 4.82 and an LLM-judged consistency score of 4.21. These scores substantially exceed those of Trace SFT at 4.31 and 3.84, as well as Serial Cascade at 4.15 and 3.67, indicating less repetitive and more logically coherent factor trajectories.

\subsection{Ablation Study}

Table~\ref{tab:ablation} isolates the contribution of each component in our method.
Ablating Value-Guided Trajectory Filtering causes the largest degradation. User Emotion M-F1 drops by 1.45 points and System Strategy M-F1 falls by 2.50 points, confirming VGTF as the primary engine for structural reasoning. Furthermore, trajectory return $J_8$ decreases by 0.66, and response generation suffers with ROUGE-L dropping by 0.73.
Trace-Conditioned Relative Optimization exclusively refines linguistic realization. Its removal leaves factor metrics unchanged but degrades generation quality, reducing ROUGE-L by 0.90 and BLEU-2 by 0.38. This proves VGTF and TCRO serve orthogonal functions regarding structural planning versus linguistic optimization.
Without explicit access to prior commitments, $J_8$ drops by 0.43 and factor accuracies moderately decline, while response metrics remain stable. This confirms the specific role of retention in stabilizing multi-turn trajectories against error propagation.
Bypassing intermediate factor prediction for direct response generation causes catastrophic collapse across all metrics, decreasing ROUGE-L by 4.71 and $J_8$ by 1.72. This validates the factorized trace as the indispensable structural backbone of our method.

\subsection{Case Study}

\begin{figure}[t]
    \centering
    \includegraphics[width=0.85\columnwidth]{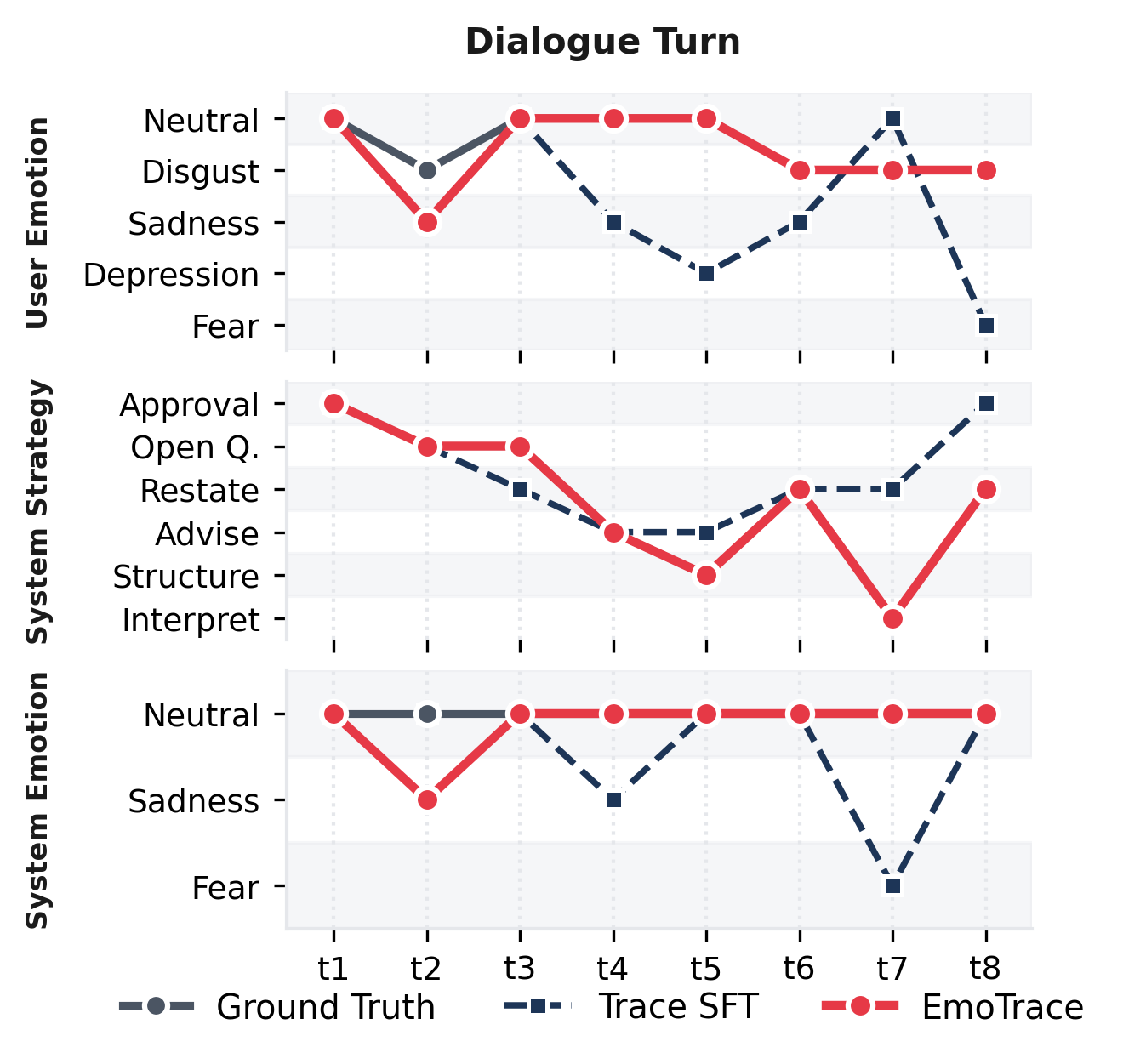}
    \caption{Per-factor trajectories over eight supporter turns. EmoTrace maintains alignment with ground truth and recovers from early mismatches like the one at increment 2, whereas Trace SFT suffers from persistent fluctuations.}
    \label{fig:case_trajectory}
\end{figure}

Figure~\ref{fig:case_trajectory} visualizes the per-factor trajectory of a held-out MESC interaction. While Trace SFT fluctuates erratically and fails to recover from early mismatches, EmoTrace maintains close alignment with the ground truth. Notably, at increment 2, EmoTrace over-reads user emotion as Sadness instead of Disgust, yet retention-anchored adjustments across all factors rejoin the optimal path at increment 3. This illustrates that EmoTrace enables globally coordinated trajectory corrections rather than myopic, independent local predictions.

\section{Conclusion}

EmoTrace reconceptualizes conversational support as a continuous evolution of structured commitments, factorizing each turn into explicit increments that are recursively retained alongside surface text. This decoupling separates extended trajectory planning from immediate linguistic realization, anchoring subsequent generations in established states and preventing the myopic drift of standard autoregressive generation. Bounded local improvements thereby compound into structural benefits over extended horizons, yielding reliable dialogue agents without external planners.

\bibliography{aaai2027}

@inproceedings{cheng2022multiesc,
  author    = {Cheng, Yi and Liu, Wenge and Li, Wenjie and Wang, Jiashuo and Zhao, Ruihui and Liu, Bang and Liang, Xiaodan and Zheng, Yefeng},
  title     = {Improving Multi-turn Emotional Support Dialogue Generation with Lookahead Strategy Planning},
  booktitle = {Proceedings of the 2022 Conference on Empirical Methods in Natural Language Processing},
  year      = {2022},
  pages     = {3014--3026},
  doi       = {10.18653/v1/2022.emnlp-main.195}
}

@inproceedings{cheng2023pal,
  author    = {Cheng, Jiale and Sabour, Sahand and Sun, Hao and Chen, Zhuang and Huang, Minlie},
  title     = {{PAL}: Persona-Augmented Emotional Support Conversation Generation},
  booktitle = {Findings of the Association for Computational Linguistics: ACL 2023},
  year      = {2023},
  pages     = {535--554},
  doi       = {10.18653/v1/2023.findings-acl.34}
}

@article{chu2025mesc,
  author  = {Chu, Yuqi and Liao, Lizi and Zhou, Zhiyuan and Ngo, Chong-Wah and Hong, Richang},
  title   = {Towards Multimodal Emotional Support Conversation Systems},
  journal = {IEEE Transactions on Multimedia},
  year    = {2025},
  volume  = {27},
  pages   = {8276--8287},
  doi     = {10.1109/TMM.2025.3604951}
}

@inproceedings{hu2022lora,
  author    = {Hu, Edward J. and Shen, Yelong and Wallis, Phillip and Allen-Zhu, Zeyuan and Li, Yuanzhi and Wang, Shean and Wang, Lu and Chen, Weizhu},
  title     = {{LoRA}: Low-Rank Adaptation of Large Language Models},
  booktitle = {International Conference on Learning Representations},
  year      = {2022},
  eprint    = {2106.09685},
  archiveprefix = {arXiv}
}

@article{li2026esca,
  author  = {Li, Jing and Luo, Yanxin and Han, Donghong and Zhan, Yimeng and Fu, Xiaoming and Qiao, Baiyou and Wu, Gang},
  title   = {{ESCA}: An Emotional Support Conversation Agent for Enhancing Reasonable Strategy Planning and Effective Expression},
  journal = {Proceedings of the AAAI Conference on Artificial Intelligence},
  year    = {2026},
  volume  = {40},
  number  = {21},
  pages   = {17526--17534},
  doi     = {10.1609/aaai.v40i21.38807}
}

@inproceedings{lin2004rouge,
  author    = {Lin, Chin-Yew},
  title     = {{ROUGE}: A Package for Automatic Evaluation of Summaries},
  booktitle = {Text Summarization Branches Out},
  year      = {2004},
  pages     = {74--81}
}

@inproceedings{lin2019moel,
  author    = {Lin, Zhaojiang and Madotto, Andrea and Shin, Jamin and Xu, Peng and Fung, Pascale},
  title     = {{MoEL}: Mixture of Empathetic Listeners},
  booktitle = {Proceedings of EMNLP-IJCNLP},
  year      = {2019},
  pages     = {121--132},
  doi       = {10.18653/v1/D19-1012}
}

@inproceedings{liu2021esconv,
  author    = {Liu, Siyang and Zheng, Chujie and Demasi, Orianna and Sabour, Sahand and Li, Yu and Yu, Zhou and Jiang, Yong and Huang, Minlie},
  title     = {Towards Emotional Support Dialog Systems},
  booktitle = {Proceedings of ACL-IJCNLP},
  year      = {2021},
  pages     = {3469--3483},
  doi       = {10.18653/v1/2021.acl-long.269}
}

@article{liu2025dynamicstrategy,
  author  = {Liu, Yiting and Li, Liang and Tu, Yunbin and Zhang, Beichen and Zha, Zheng-Jun and Huang, Qingming},
  title   = {Dynamic Strategy Prompt Reasoning for Emotional Support Conversation},
  journal = {IEEE Transactions on Multimedia},
  year    = {2025},
  volume  = {27},
  pages   = {108--119},
  doi     = {10.1109/TMM.2024.3521669}
}

@inproceedings{majumder2020mime,
  author    = {Majumder, Navonil and Hong, Pengfei and Peng, Shanshan and Lu, Jiankun and Ghosal, Deepanway and Gelbukh, Alexander and Mihalcea, Rada and Poria, Soujanya},
  title     = {{MIME}: Mimicking Emotions for Empathetic Response Generation},
  booktitle = {Proceedings of EMNLP},
  year      = {2020},
  pages     = {8968--8979},
  doi       = {10.18653/v1/2020.emnlp-main.721}
}

@inproceedings{papineni2002bleu,
  author    = {Papineni, Kishore and Roukos, Salim and Ward, Todd and Zhu, Wei-Jing},
  title     = {{BLEU}: A Method for Automatic Evaluation of Machine Translation},
  booktitle = {Proceedings of the 40th Annual Meeting of the Association for Computational Linguistics},
  year      = {2002},
  pages     = {311--318},
  doi       = {10.3115/1073083.1073135}
}

@inproceedings{rashkin2019empathetic,
  author    = {Rashkin, Hannah and Smith, Eric Michael and Li, Margaret and Boureau, Y-Lan},
  title     = {Towards Empathetic Open-domain Conversation Models: A New Benchmark and Dataset},
  booktitle = {Proceedings of the 57th Annual Meeting of the Association for Computational Linguistics},
  year      = {2019},
  pages     = {5370--5381},
  doi       = {10.18653/v1/P19-1534}
}

@inproceedings{roller2021blenderbot,
  author    = {Roller, Stephen and Dinan, Emily and Goyal, Naman and Ju, Da and Williamson, Mary and Liu, Yinhan and Xu, Jing and Ott, Myle and Smith, Eric Michael and Boureau, Y-Lan and Weston, Jason},
  title     = {Recipes for Building an Open-Domain Chatbot},
  booktitle = {Proceedings of the 16th Conference of the European Chapter of the Association for Computational Linguistics},
  year      = {2021},
  pages     = {300--325},
  doi       = {10.18653/v1/2021.eacl-main.24}
}

@article{sabour2022cem,
  author  = {Sabour, Sahand and Zheng, Chujie and Huang, Minlie},
  title   = {{CEM}: Commonsense-Aware Empathetic Response Generation},
  journal = {Proceedings of the AAAI Conference on Artificial Intelligence},
  year    = {2022},
  volume  = {36},
  number  = {10},
  pages   = {11229--11237},
  doi     = {10.1609/aaai.v36i10.21373}
}

@inproceedings{tu2022misc,
  author    = {Tu, Quan and Li, Yanran and Cui, Jianwei and Wang, Bin and Wen, Ji-Rong and Yan, Rui},
  title     = {{MISC}: A Mixed Strategy-Aware Model Integrating {COMET} for Emotional Support Conversation},
  booktitle = {Proceedings of the 60th Annual Meeting of the Association for Computational Linguistics},
  year      = {2022},
  pages     = {308--319},
  doi       = {10.18653/v1/2022.acl-long.25}
}

@inproceedings{wang2025straq,
  author    = {Wang, Xiaoyu and Zhao, Yue and Gu, Qingqing and Jiang, Zhonglin and Chen, Yong and Ji, Luo},
  title     = {Convert Language Model into a Value-Based Strategic Planner},
  booktitle = {Proceedings of the 63rd Annual Meeting of the Association for Computational Linguistics: Industry Track},
  year      = {2025},
  pages     = {1444--1456},
  doi       = {10.18653/v1/2025.acl-industry.102}
}

@article{schulman2017ppo,
  author        = {Schulman, John and Wolski, Filip and Dhariwal, Prafulla and Radford, Alec and Klimov, Oleg},
  title         = {Proximal Policy Optimization Algorithms},
  journal       = {arXiv preprint arXiv:1707.06347},
  year          = {2017},
  eprint        = {1707.06347},
  archiveprefix = {arXiv}
}

@inproceedings{rafailov2023dpo,
  author    = {Rafailov, Rafael and Sharma, Archit and Mitchell, Eric and Manning, Christopher D. and Ermon, Stefano and Finn, Chelsea},
  title     = {Direct Preference Optimization: Your Language Model Is Secretly a Reward Model},
  booktitle = {Advances in Neural Information Processing Systems},
  year      = {2023},
  volume    = {36},
  pages     = {53728--53741},
  publisher = {Curran Associates, Inc.}
}

@article{shao2024deepseekmath,
  author        = {Shao, Zhihong and Wang, Peiyi and Zhu, Qihao and Xu, Runxin and Song, Junxiao and Bi, Xiao and Zhang, Haowei and Zhang, Mingchuan and Li, Y. K. and Wu, Y. and Guo, Daya},
  title         = {{DeepSeekMath}: Pushing the Limits of Mathematical Reasoning in Open Language Models},
  journal       = {arXiv preprint arXiv:2402.03300},
  year          = {2024},
  eprint        = {2402.03300},
  archiveprefix = {arXiv}
}

@article{yang2024qwen25,
  author  = {Yang, An and Yang, Baosong and Zhang, Beichen and Hui, Binyuan and Zheng, Bo and Yu, Bowen and others},
  title   = {{Qwen2.5} Technical Report},
  journal = {arXiv preprint arXiv:2412.15115},
  year    = {2024},
  eprint  = {2412.15115},
  archiveprefix = {arXiv},
  doi     = {10.48550/arXiv.2412.15115},
  url     = {https://arxiv.org/abs/2412.15115}
}

@inproceedings{zhang2020bertscore,
  author    = {Zhang, Tianyi and Kishore, Varsha and Wu, Felix and Weinberger, Kilian Q. and Artzi, Yoav},
  title     = {{BERTScore}: Evaluating Text Generation with {BERT}},
  booktitle = {International Conference on Learning Representations},
  year      = {2020},
  eprint    = {1904.09675},
  archiveprefix = {arXiv}
}

@inproceedings{zhang2025decoupledesc,
  author    = {Zhang, Chao and Shi, Xin and Zhang, Xueqiao and Zhu, Yifan and Yang, Yi and Luo, Yawei},
  title     = {{DecoupledESC}: Enhancing Emotional Support Generation via Strategy-Response Decoupled Preference Optimization},
  booktitle = {Findings of the Association for Computational Linguistics: EMNLP 2025},
  year      = {2025},
  pages     = {22189--22215},
  doi       = {10.18653/v1/2025.findings-emnlp.1209}
}

@inproceedings{zhao2023transesc,
  author    = {Zhao, Weixiang and Zhao, Yanyan and Wang, Shilong and Qin, Bing},
  title     = {{TransESC}: Smoothing Emotional Support Conversation via Turn-Level State Transition},
  booktitle = {Findings of the Association for Computational Linguistics: ACL 2023},
  year      = {2023},
  pages     = {6725--6739},
  doi       = {10.18653/v1/2023.findings-acl.420}
}

@article{ma2020survey,
  author  = {Ma, Yukun and Nguyen, Khanh Linh and Xing, Frank Z. and Cambria, Erik},
  title   = {A Survey on Empathetic Dialogue Systems},
  journal = {Information Fusion},
  year    = {2020},
  volume  = {64},
  pages   = {50--70},
  doi     = {10.1016/j.inffus.2020.06.011}
}

@article{reeck2016socialregulation,
  author  = {Reeck, Crystal and Ames, Daniel R. and Ochsner, Kevin N.},
  title   = {The Social Regulation of Emotion: An Integrative, Cross-Disciplinary Model},
  journal = {Trends in Cognitive Sciences},
  year    = {2016},
  volume  = {20},
  number  = {1},
  pages   = {47--63},
  doi     = {10.1016/j.tics.2015.09.003}
}

@inproceedings{deng2023kemi,
  author    = {Deng, Yang and Zhang, Wenxuan and Yuan, Yifei and Lam, Wai},
  title     = {Knowledge-Enhanced Mixed-Initiative Dialogue System for Emotional Support Conversations},
  booktitle = {Proceedings of the 61st Annual Meeting of the Association for Computational Linguistics},
  year      = {2023},
  pages     = {4079--4095},
  doi       = {10.18653/v1/2023.acl-long.225},
  url       = {https://aclanthology.org/2023.acl-long.225/}
}

@inproceedings{zhou2023supporter,
  author    = {Zhou, Jinfeng and Chen, Zhuang and Wang, Bo and Huang, Minlie},
  title     = {Facilitating Multi-Turn Emotional Support Conversation with Positive Emotion Elicitation: A Reinforcement Learning Approach},
  booktitle = {Proceedings of the 61st Annual Meeting of the Association for Computational Linguistics},
  year      = {2023},
  pages     = {1714--1729},
  doi       = {10.18653/v1/2023.acl-long.96},
  url       = {https://aclanthology.org/2023.acl-long.96/}
}

@article{peng2023fado,
  author  = {Peng, Wei and Qin, Ziyuan and Hu, Yue and Xie, Yuqiang and Li, Yunpeng},
  title   = {{FADO}: Feedback-Aware Double Controlling Network for Emotional Support Conversation},
  journal = {Knowledge-Based Systems},
  year    = {2023},
  volume  = {264},
  pages   = {110340},
  doi     = {10.1016/j.knosys.2023.110340}
}

@inproceedings{liu2022prophetchat,
  author    = {Liu, Chang and Tan, Xu and Tao, Chongyang and Fu, Zhenxin and Zhao, Dongyan and Liu, Tie-Yan and Yan, Rui},
  title     = {{ProphetChat}: Enhancing Dialogue Generation with Simulation of Future Conversation},
  booktitle = {Proceedings of the 60th Annual Meeting of the Association for Computational Linguistics},
  year      = {2022},
  pages     = {962--973},
  doi       = {10.18653/v1/2022.acl-long.68},
  url       = {https://aclanthology.org/2022.acl-long.68/}
}

@article{li2022kemp,
  author    = {Li, Qintong and Li, Piji and Ren, Zhaochun and Ren, Pengjie and Chen, Zhumin},
  title     = {Knowledge Bridging for Empathetic Dialogue Generation},
  journal   = {Proceedings of the AAAI Conference on Artificial Intelligence},
  year      = {2022},
  volume    = {36},
  number    = {10},
  pages     = {10993--11001},
  doi       = {10.1609/aaai.v36i10.21347}
}

@inproceedings{zhou2023case,
  author    = {Zhou, Jinfeng and Zheng, Chujie and Wang, Bo and Zhang, Zheng and Huang, Minlie},
  title     = {{CASE}: Aligning Coarse-to-Fine Cognition and Affection for Empathetic Response Generation},
  booktitle = {Proceedings of the 61st Annual Meeting of the Association for Computational Linguistics},
  year      = {2023},
  pages     = {8223--8237},
  doi       = {10.18653/v1/2023.acl-long.457},
  url       = {https://aclanthology.org/2023.acl-long.457/}
}

@inproceedings{zhao2018discrete,
  author    = {Zhao, Tiancheng and Lee, Kyusong and Eskenazi, Maxine},
  title     = {Unsupervised Discrete Sentence Representation Learning for Interpretable Neural Dialog Generation},
  booktitle = {Proceedings of the 56th Annual Meeting of the Association for Computational Linguistics},
  year      = {2018},
  pages     = {1098--1107},
  doi       = {10.18653/v1/P18-1101},
  url       = {https://aclanthology.org/P18-1101/}
}

@inproceedings{bao2020plato,
  author    = {Bao, Siqi and He, Huang and Wang, Fan and Wu, Hua and Wang, Haifeng},
  title     = {{PLATO}: Pre-Trained Dialogue Generation Model with Discrete Latent Variable},
  booktitle = {Proceedings of the 58th Annual Meeting of the Association for Computational Linguistics},
  year      = {2020},
  pages     = {85--96},
  doi       = {10.18653/v1/2020.acl-main.9},
  url       = {https://aclanthology.org/2020.acl-main.9/}
}

@article{he2022galaxy,
  author    = {He, Wanwei and Dai, Yinpei and Zheng, Yinhe and Wu, Yuchuan and Cao, Zheng and Liu, Dermot and Jiang, Peng and Yang, Min and Huang, Fei and Si, Luo and Sun, Jian and Li, Yongbin},
  title     = {{GALAXY}: A Generative Pre-Trained Model for Task-Oriented Dialog with Semi-Supervised Learning and Explicit Policy Injection},
  journal   = {Proceedings of the AAAI Conference on Artificial Intelligence},
  year      = {2022},
  volume    = {36},
  number    = {10},
  pages     = {10749--10757},
  doi       = {10.1609/aaai.v36i10.21320}
}

@inproceedings{wang2024muffin,
  author    = {Wang, Jiashuo and Xu, Chunpu and Leong, Chak Tou and Li, Wenjie and Li, Jing},
  title     = {{Muffin}: Mitigating Unhelpfulness in Emotional Support Conversations with Multifaceted {AI} Feedback},
  booktitle = {Findings of the Association for Computational Linguistics: ACL 2024},
  year      = {2024},
  pages     = {567--585},
  doi       = {10.18653/v1/2024.findings-acl.31},
  url       = {https://aclanthology.org/2024.findings-acl.31/}
}

@inproceedings{kang2024preference,
  author    = {Kang, Dongjin and Kim, Sunghwan and Kwon, Taeyoon and Moon, Seungjun and Cho, Hyunsouk and Yu, Youngjae and Lee, Dongha and Yeo, Jinyoung},
  title     = {Can Large Language Models Be Good Emotional Supporter? Mitigating Preference Bias on Emotional Support Conversation},
  booktitle = {Proceedings of the 62nd Annual Meeting of the Association for Computational Linguistics},
  year      = {2024},
  pages     = {15232--15261},
  doi       = {10.18653/v1/2024.acl-long.813},
  url       = {https://aclanthology.org/2024.acl-long.813/}
}

@inproceedings{zhong2020persona,
  author    = {Zhong, Peixiang and Zhang, Chen and Wang, Hao and Liu, Yong and Miao, Chunyan},
  title     = {Towards Persona-Based Empathetic Conversational Models},
  booktitle = {Proceedings of the 2020 Conference on Empirical Methods in Natural Language Processing},
  year      = {2020},
  pages     = {6556--6566},
  doi       = {10.18653/v1/2020.emnlp-main.531},
  url       = {https://aclanthology.org/2020.emnlp-main.531/}
}

@inproceedings{zhao2019latentactions,
  author    = {Zhao, Tiancheng and Xie, Kaige and Eskenazi, Maxine},
  title     = {Rethinking Action Spaces for Reinforcement Learning in End-to-End Dialog Agents with Latent Variable Models},
  booktitle = {Proceedings of NAACL-HLT},
  year      = {2019},
  pages     = {1208--1218},
  doi       = {10.18653/v1/N19-1123},
  url       = {https://aclanthology.org/N19-1123/}
}

@article{li2019dialogueirl,
  author    = {Li, Ziming and Kiseleva, Julia and de Rijke, Maarten},
  title     = {Dialogue Generation: From Imitation Learning to Inverse Reinforcement Learning},
  journal   = {Proceedings of the AAAI Conference on Artificial Intelligence},
  year      = {2019},
  volume    = {33},
  number    = {1},
  pages     = {6722--6729},
  doi       = {10.1609/aaai.v33i01.33016722}
}

@inproceedings{christiano2017humanpreferences,
  author    = {Christiano, Paul F. and Leike, Jan and Brown, Tom B. and Martic, Miljan and Legg, Shane and Amodei, Dario},
  title     = {Deep Reinforcement Learning from Human Preferences},
  booktitle = {Advances in Neural Information Processing Systems},
  year      = {2017},
  volume    = {30},
  pages     = {4299--4307},
  eprint    = {1706.03741},
  archiveprefix = {arXiv}
}

@inproceedings{stiennon2020summarize,
  author    = {Stiennon, Nisan and Ouyang, Long and Wu, Jeffrey and Ziegler, Daniel M. and Lowe, Ryan and Voss, Chelsea and Radford, Alec and Amodei, Dario and Christiano, Paul F.},
  title     = {Learning to Summarize with Human Feedback},
  booktitle = {Advances in Neural Information Processing Systems},
  year      = {2020},
  volume    = {33},
  pages     = {3008--3021},
  eprint    = {2009.01325},
  archiveprefix = {arXiv}
}

@article{ziegler2019finetuning,
  author  = {Ziegler, Daniel M. and Stiennon, Nisan and Wu, Jeffrey and Brown, Tom B. and Radford, Alec and Amodei, Dario and Christiano, Paul and Irving, Geoffrey},
  title   = {Fine-Tuning Language Models from Human Preferences},
  journal = {arXiv preprint arXiv:1909.08593},
  year    = {2019},
  eprint  = {1909.08593},
  archiveprefix = {arXiv},
  doi     = {10.48550/arXiv.1909.08593},
  url     = {https://arxiv.org/abs/1909.08593}
}

@inproceedings{ghosal2019dialoguegcn,
  author    = {Ghosal, Deepanway and Majumder, Navonil and Poria, Soujanya and Chhaya, Niyati and Gelbukh, Alexander},
  title     = {{DialogueGCN}: A Graph Convolutional Neural Network for Emotion Recognition in Conversation},
  booktitle = {Proceedings of EMNLP-IJCNLP},
  year      = {2019},
  pages     = {154--164},
  doi       = {10.18653/v1/D19-1015},
  url       = {https://aclanthology.org/D19-1015/}
}

@inproceedings{hu2021mmgcn,
  author    = {Hu, Jingwen and Liu, Yuchen and Zhao, Jinming and Jin, Qin},
  title     = {{MMGCN}: Multimodal Fusion via Deep Graph Convolution Network for Emotion Recognition in Conversation},
  booktitle = {Proceedings of ACL-IJCNLP},
  year      = {2021},
  pages     = {5666--5675},
  doi       = {10.18653/v1/2021.acl-long.440},
  url       = {https://aclanthology.org/2021.acl-long.440/}
}

@inproceedings{hu2022mmdfn,
  author    = {Hu, Dou and Hou, Xiaolong and Wei, Lingwei and Jiang, Lianxin and Mo, Yang},
  title     = {{MM-DFN}: Multimodal Dynamic Fusion Network for Emotion Recognition in Conversations},
  booktitle = {Proceedings of ICASSP},
  year      = {2022},
  pages     = {7037--7041},
  doi       = {10.1109/ICASSP43922.2022.9747397}
}

@article{openai2023gpt4,
  author  = {{OpenAI}},
  title   = {{GPT}-4 Technical Report},
  journal = {arXiv preprint arXiv:2303.08774},
  year    = {2023},
  eprint  = {2303.08774},
  archiveprefix = {arXiv},
  doi     = {10.48550/arXiv.2303.08774},
  url     = {https://arxiv.org/abs/2303.08774}
}

@article{grattafiori2024llama3,
  author  = {Grattafiori, Aaron and others},
  title   = {The Llama 3 Herd of Models},
  journal = {arXiv preprint arXiv:2407.21783},
  year    = {2024},
  eprint  = {2407.21783},
  archiveprefix = {arXiv},
  doi     = {10.48550/arXiv.2407.21783},
  url     = {https://arxiv.org/abs/2407.21783}
}


\end{document}